\documentclass{article}

\usepackage{microtype}
\usepackage{graphicx}
\usepackage{subfigure}
\usepackage{booktabs} 
\usepackage{bm}
\usepackage{hyperref}

\usepackage{etoolbox}
\makeatletter
\patchcmd{\Hy@Warning}{empty anchor}{empty anchor ignored}{}{}
\makeatother

\usepackage{makecell}
\usepackage{diagbox}
\usepackage{multirow}
\usepackage{threeparttable}

\usepackage{hyperref}
\hypersetup{
colorlinks=true,
linkcolor=blue,
anchorcolor=blue,
citecolor=blue}

\usepackage[accepted]{icml2024}

\usepackage{amsmath}
\usepackage{amssymb}
\usepackage{mathtools}
\usepackage{amsthm}

\usepackage[capitalize,noabbrev]{cleveref}

\theoremstyle{plain}

\theoremstyle{definition}

\theoremstyle{remark}

\usepackage{bbm}

\newcommand*{\argmax}{\mathop{\mathrm{argmax}}}
\usepackage[textsize=tiny]{todonotes}

\icmltitlerunning{Neuro-Symbolic Temporal Point Processes}

\begin{document}

\twocolumn[
\icmltitle{Neuro-Symbolic Temporal Point Processes}



\icmlsetsymbol{equal}{*}

\begin{icmlauthorlist}
\icmlauthor{Yang Yang}{sch}
\icmlauthor{Chao Yang}{sch}
\icmlauthor{Boyang Li}{sch}
\icmlauthor{Yinghao Fu}{sch}
\icmlauthor{Shuang Li}{sch}
\end{icmlauthorlist}

\icmlaffiliation{sch}{School of Data Science, The Chinese University of Hong Kong (Shenzhen)}

\icmlcorrespondingauthor{Shuang Li}{lishuang@cuhk.edu.cn}

\icmlkeywords{Machine Learning, ICML}

\vskip 0.3in
]



\printAffiliationsAndNotice{}  

\begin{abstract}
Our goal is to {\it efficiently} discover a compact set of temporal logic rules to explain irregular events of interest. We introduce a neural-symbolic rule induction framework within the temporal point process model. The negative log-likelihood is the loss that guides the learning, where the explanatory logic rules and their weights are learned end-to-end in a {\it differentiable} way. Specifically, predicates and logic rules are represented as {\it vector embeddings}, where the predicate embeddings are fixed and the rule embeddings are trained via gradient descent to obtain the most appropriate compositional representations of the predicate embeddings. To make the rule learning process more efficient and flexible, we adopt a {\it sequential covering algorithm}, which progressively adds rules to the model and removes the event sequences that have been explained until all event sequences have been covered. All the found rules will be fed back to the models for a final rule embedding and weight refinement. 
Our approach showcases notable efficiency and accuracy across synthetic and real datasets, surpassing state-of-the-art baselines by a wide margin in terms of efficiency.
\end{abstract}

\section{Introduction}
\label{sec:intr}
Explaining critical events, such as sudden health changes or unusual transactions, is essential in high-stakes domains like healthcare and finance. The dynamics of these events are typically governed by temporal logic rules, and automatically uncovering these rules from data holds significant scientific and practical value.

For example, in healthcare, it is desirable to compress and summarize medical knowledge or clinical experiences regarding disease phenotypes and therapies to a collection of temporal logic rules. The discovered rules can contribute to the sharing of clinical experiences and aid in the improvement of the treatment strategy. They can also provide specific explanations for the occurrence of an event. For example, the following clinical report 

``{\it A 50-year-old patient, with a chronic lung disease since 5 years ago, took the booster vaccine shot on March 1st. The patient got exposed to the COVID-19 virus around May 12th, and afterward within a week began to have a mild cough and nasal congestion. The patient received treatment as soon as the symptoms appeared. After intravenous infusions at a healthcare facility for around 3 consecutive days, the patient recovered...}''

contains many clinical events with timestamps recorded. It sounds appealing to distill compact human-readable temporal logic rules from these noisy event data to aid diagnoses and treatment planning. In this paper, we present an {\it efficient} neural-symbolic rule induction algorithm capable of automatically learning {\it universal} rules from sequences of irregular event data. These universal rules act as summarized laws that effectively elucidate the dynamics of the events, offering valuable insights for clinical decision-making.

From modeling perspective, we design a {\it neural-symbolic temporal point process} (NS-TPP) that strikes the balance between model flexibility and interpretability. The occurrence rate (i.e., intensity) of events is a function of the neural predicate embeddings, where the functional form is determined by the logic rules that are uncovered from the data.  Traditional parametric temporal point process (TPP) models like the Hawkes process offer interpretability, but their simplicity limits flexibility. Conversely, neural-based models, such as RMTPP~\cite{du2016recurrent} and Transformer Hawkes~\cite{zuo2020transformer}, provide expressiveness but are often criticized for their black-box nature and hinder their applications in high-stakes scenarios. Our NS-TPP strives to harness the strengths of both paradigms.

To enable {\it efficient} and {\it differentiable} rule learning, we propose a {\it neural-symbolic rule induction} framework for TPP, which aims to learn rule embeddings to identify the rule formula. In our model, predicates, or logic variables, are represented as fixed vector embeddings, either pre-trained or specified beforehand. Each rule embedding acts as a {\it learnable filter}, selecting the most relevant predicates and evidence from observational facts to form logical rules. During the forward pass, these filters scan predicate embeddings to find the best matches and combined with the observed events as fact, these filters generate logic-informed features. This forward pass can be thought of as using the rule content filter on the historical events to gather evidence, which will then be used to deduce the occurrence of the event of interest. In the backward pass, we calculate the loss as the negative log-likelihood based on a temporal point process. The rule embedding parameters are then optimized end-to-end through gradient descent. 

Furthermore, to boost {\it flexibility} in rule learning, we utilize a {\it sequential covering} algorithm. This method involves progressively adding rules to the model and learning each rule embedding one by one. When a new rule is identified (i.e., the learning of the new rule embedding converges), the event sequences it explains are removed from the dataset. This rule learning process continues until all events are covered, which naturally eliminates the need to specify the total number of rules in advance. After identifying all rule embeddings and their weights using the sequential covering algorithm, we jointly refine the rule embeddings and weights by considering the full NS-TPP model. In this way, we further enhance the accuracy of rule embeddings and weights by maximizing the likelihood.

In summary, our contributions are as follows:

({\it i}) Our NS-TPP model incorporates a neural-symbolic intensity function, striking a balance between flexibility and interpretability. By converting model structure learning into rule embedding learning, the discovered rule set automatically determines the model capacity and structure. All the rule embeddings and other model parameters will be learned in a differentiable way.

({\it ii}) Our neural-symbolic rule induction algorithm naturally withstands input noise. This resilience is achieved through encoding rule content and predicates using embeddings. By computing features based on similarity scores among rule embedding, predicate embedding, and relevant facts, our approach ensures robustness to noisy inputs.

({\it iii}) We improve rule discovery efficiency and flexibility by implementing a covering algorithm, dividing the complete learning problem into manageable sub-problems. Our algorithm's efficiency and accuracy are validated on both synthetic and real data, demonstrating approximately 100 times greater efficiency.

\section{Related Work}
We will compare our method with some existing works from the following aspects.
\vspace{-8pt}
\paragraph{Temporal Point Process (TPP) Models} TPP models have emerged as an elegant framework for modeling event times and types in continuous time, directly treating the inter-event times as random variables. Advances in this field have largely concentrated on enhancing the flexibility of intensity functions to improve event prediction accuracy. Pioneering works such as the RMTPP  \cite{du2016recurrent} and the continuous-time RNN further improved from RMTPP \cite{mei2017neural} introduced recurrent neural network-based approaches to model the intensity functions. More recent studies by \citet{zuo2020transformer} and \citet{zhang2020self} have applied the self-attention mechanism to address long-term event dependencies, showcasing the potential of leveraging attention-based deep learning techniques for TPP. Despite these advancements, the reliance on black-box models raises significant interpretability issues, particularly in contexts requiring  explanations for events, such as root cause analysis for abnormal events. This gap highlights the increasing agreement on the need for inherently interpretable models, as emphasized by \citet{rudin2019stop}, to ensure the transparency of the decision-making in high-stakes systems.

In response to these challenges, \citet{li2020temporal,li2021explaining} proposed integrating logic rules within the intensity function to foster interpretability. However, their methods either assume that the logic rules are prespecified or rely on a non-differentiable rule learning process, distinguishing our approach which offers a differentiable framework for rule learning.


\paragraph{Rule Mining} Discovering rules from data in an unsupervised manner has long been a challenging task. Unsupervised logic rule mining is about discovering inherent patterns in data without any prior labeling. Traditional approaches, such as Itemset Mining Methods like Apriori \cite{agrawal1994fast} and NEclatclosed \cite{aryabarzan2021neclatclosed}, focus on identifying frequent itemsets. However, they cannot be directly adapted to events with recorded occurrence times, limiting their applicability in temporal datasets. On the other hand, Sequential Pattern Mining methods like CM-SPADE \cite{fournier2014fast} and VGEN \cite{fournier2014vgen} aim to uncover temporal relationships in datasets. However, they only utilize the temporal ordering of events and are unable to effectively incorporate fine-grained timestamp information, which can lead to precision issues in rule mining.

Supervised logic rule mining requires labeled data, consisting of both positive and negative samples. The rules are mined usually under the principle that, for positive samples, at least one rule must be satisfied, and for negative samples, none of the rules should be satisfied. Among supervised rule mining methods, a notable example is Inductive Logic Programming (ILP) \cite{srinivasan2001aleph}, which provides a structured framework for rule learning. However, ILP typically requires a balanced mix of positive and negative examples to achieve effective rule learning. The ILP methods can be categorized into forward-chaining methods \cite{Campero2018LogicalRI,payani2019inductive}, which generate and test rules through iterative deductive reasoning, and backward-chaining methods \cite{minervini2018towards,yang2022temporal}, which dynamically construct rules to satisfy specific queries or goals. These approaches, despite their innovative attempts at rule induction, often operate as opaque models. They lack the ability to clearly explain the reasoning behind their inferences, making them more like black boxes than interpretable systems.

Our NS-TPP learns temporal logic rules from data in an unsupervised manner, utilizing fine-grained temporal information without requiring positive or negative labeling. This extends unsupervised temporal logic rule discovery methods, broadening the scope of rule learning without relying on labeled data.


\section{Background}
\paragraph{Predicates and Temporal Logic Rules}
Define a set of predicates as $\mathcal{X}$, where each variable $X_u \in \mathcal{X}$ is a boolean logic variable. Denote the target predicate we aim to explain as $Y \in \mathcal{X}$. For example, $Y$ could represent a sudden change in a patient's health, an unusually large transaction, or an alarm in manufacturing. We assume that the target predicate can be explained by a set of Horn rules (i.e., if-then rules) with temporal ordering constraints, each having the general form:{\small \begin{align}
f: \quad Y \leftarrow\left(\bigwedge_{X_u \in \mathcal{X}_f} X_u\right) \bigwedge\left(\bigwedge_{X_u, X_v \in \mathcal{X}_f} R\left(X_u, X_v\right)\right) 
\label{eq:rule}
\end{align}}
where $\mathcal{X}_f$ is the set of body predicates associated with rule $f$, and $R\left(X_u, X_v\right)$ represents temporal relations between each paired predicates $X_u$ and $X_v$. These relations, categorized as ``Before", ``Equal", ``After" or ``None", define the temporal constraints between $X_u$ and $X_v$, with ``None" specifying the absence of any temporal relation.  

\paragraph{Temporal Point Process (TPP)} Consider adding a temporal dimension to the previously defined static predicates, and the grounded predicates by observed data (i.e., fact) results in a list of spiked events, denoted as $\left\{X_u(t)\right\}_{t \geq 0}$, where each $X_u(t) \in\{0,1\}$ at any time $t\geq0$. Specifically, $X_u(t)$ transitions instantaneously from 0 (False) to 1 (True) at the timestamp when the event occurs. In our context, each event sequence sample represents a $|\mathcal{X}|$-dimensional multivariate temporal point process. We use $\mathcal{H}_{t^-}=\left\{X_u(t)\right\}_{u=1, \ldots,|\mathcal{X}|}$ to denote all the observed events up to but not including $t$.

We are interested in modeling and learning logical explanations to the occurrence of the target event sequence $\{Y(t)\}_{t \geq 0}$ with event time recorded as $\{t_1, t_2, \dots\}$. We treat the inter-event time intervals as random variables and the duration until the next event $Y$ is characterized by the conditional intensity function, denoted as $\lambda\left(t \mid \mathcal{H}_{t^-}\right)$. By definition, 
$$\lambda(t \mid\mathcal{H}_{t^-}) d t=\mathbb{E}\left[N([t, t+d t]) \mid \mathcal{H}_{t^-}\right]$$
where $N([t, t+d t])$ denotes the number of events occurring in the interval $[t, t+d t)$. Given the occurrence time of event $Y$, such as $\left(t_1, \ldots, t_n\right)$, the joint likelihood function of the data is computed by $p\left(t_1, \ldots, t_n\right)=\prod_{i=1}^n p^*\left(t_i\right)$ using the chain rule, where the conditional probability
$$
p^*\left(t_i\right)=\lambda^*\left(t_i\right) \exp \left(-\int_{t_{i-1}}^{t_i} \lambda^*(s) d s\right) .
$$
Here, to simplify the notation, we denote $p^*(t):=$ $p\left(t \mid \mathcal{H}_{t^-}\right)$ and $\lambda^*(t):=\lambda\left(t \mid \mathcal{H}_{t^-}\right)$. 

In this paper, we will model $\lambda^*(t)$ using neural-symbolic features, and we name our model as NS-TPP. Moreover, we aim to design a neural-symbolic rule induction algorithm to efficiently uncover the logic rule set $\mathcal{F}:=\left\{f_1, f_2, \ldots, f_H\right\}$ and learn other continuous model parameters jointly through maximizing the likelihood by gradient descent. Given the learned NS-TPP, we can deduce and explain the occurrence of target events in a {\it probabilistic} and {\it continuous-time} manner. 
\section{Neural-Symbolic Temporal Point Process (NS-TPP)}
\begin{figure*}[t]
    \centering
    \includegraphics[scale=0.42]{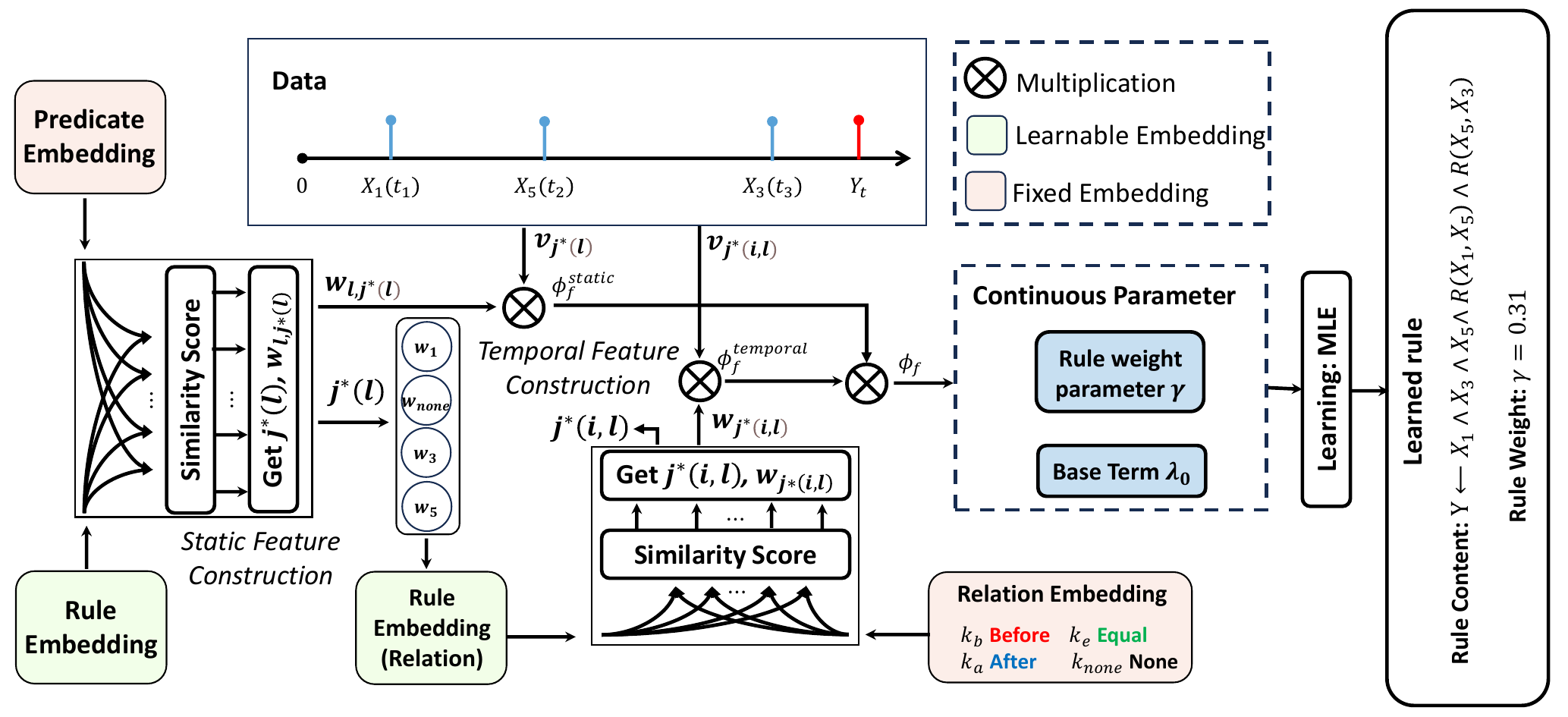}
    \vspace{-10pt}
    \caption{Overview of our neural-symbolic framework for temporal logic induction. The framework begins with preparing fixed predicate embeddings. During the forward pass, rule embeddings scan these predicate embeddings to identify optimal compositional matches. A modified attention-like block then integrates these matches with observed events to generate static neural-symbolic features. Temporal relations are then incorporated to create neural-symbolic temporal features given the already selected predicate pairs. We will use the same matching idea to obtain the temporal neural-symblic features. The final neural-symbolic features by combing static and temporal parts will be used to compute the intensity function and the likelihood. We will adopt MLE to learn all the rule embedding parameters and other continuous model parameters (rule weights and base term) in a differentiable way. The overall rule learning scheme employs the sequential covering algorithm to learn rules progressively until no further rules can be added. Finally, all the rule embeddings and other model parameters will be refined to optimize the likelihood.
    }
    \label{fig:framework}
    \label{Figure1}
\end{figure*}
\subsection{Neural-Symbolic Feature Construction}
The core idea of our proposed NS-TPP is to formulate the neural-symbolic features to construct the intensity function for $\{Y(t)\}_{t \geq 0}$. Let's temporarily assume that all rules are known, denoted as $\mathcal{F}$, and we model the intensity function as:
\begin{align}
\lambda^*\left(t  \mid \mathcal{F} \right)=b_0 +  \sum_{f \in \mathcal{F}} \gamma_f \phi_{f}\left(\mathcal{H}_{t^-}\right)
\label{eq:intensity}
\end{align}
where $b_0$ represents the base term independent of rules, $\gamma_f$ denotes the impact weight of each rule $f \in \mathcal{F}$, and $\phi_f(\cdot)$ is the neural-symbolic feature that depends on rules and data. We will elaborate on how to compute the neural-symbolic feature $\phi_{f}\left(\mathcal{H}_{t^-}\right)$ below and the overall framework is illustrated in Figure \ref{fig:framework}.
\paragraph{Predicate Embedding} For each predicate $X\in \mathcal{X}$, we represent it as a row embedding vector. All the predciate embeddings are denoted as $\bm{k}_1, \bm{k}_2, \dots, \bm{k}_{|\mathcal{X}|}$, each of dimension $d$, i.e., $\bm{k}_i \in \mathbb{R}^{1\times d}$. These predicate embeddings can be obtained through pretraining or prespecification. These embeddings can take various forms, such as one-hot representations, which are simple binary vectors, or dense vector embeddings extracted from pretrained models like neural TPP (e.g., Transformer Hawkes), which may capture the semantic dependency between predicates. We also introduce and specify a dummy predicate embedding (e.g., as a zero vector), denoted as ${\bm{k}_0}$, to signify a predicate with no semantic meaning. We will show later that the introduction of the dummy predicate embedding is to accommodate various rule lengths in rule learning.

Regardless of how the predicate embeddings are obtained, it is essential that each predicate embedding is distinct and carries concrete semantic meaning. This is crucial for interpreting the rule formula from the learned rule embeddings. We denote the stacked predicate embedding matrix as $\bm{K}=[\bm{k}_0; \bm{k}_1; \dots,  \bm{k}_{|\mathcal{X}|}] \in  \mathbb{R}^{ (|\mathcal{X}|+1) \times d}$.

\paragraph{Rule Embedding} Now, let's introduce the {\it rule embedding} that will be learned from data to indicate a rule formula $f$. Each rule embedding will act as a learnable filter to compute the similarity score with the predicate embeddings, selecting predicates to form a rule and gather evidence from data to construct the neural-symbolic feature.

Each rule embedding encodes one rule. Suppose we aim to learn a rule with length $L$, we will initialize the rule embedding as $\bm{Q}_f=[\bm{q}_1; \bm{q}_2; \dots; \bm{q}_{L}] \in  \mathbb{R}^{L \times d}$, where each row vector $\bm{q}_l \in \mathbb{R}^{1 \times d}$, sharing the same dimension with the predicate embedding, and $L$ indicates the {\it maximum} rule length. 

\paragraph{Neural-Symbolic Feature} The fundamental concept behind the proposed rule induction is that the rule embedding $\bm{Q}_f$ can be regarded as $L$ slots to be filled in by predicate embeddings. Learning the rule embedding will dynamically decide which predicates to select to form  the rule, and the selection score is based on the similarity between each current rule embedding vector and all the predicate embedding vectors. Written in matrix form, we can determine which predicate embeddings to fill in each rule embedding slot by computing the similarity score:
\begin{align}
{\bm W} =\operatorname{softmax}(\bm{Q}_f \bm{K}^{\top} /\tau)
\label{Eq.2}
\end{align}
where softmax is applied row-wise. The similarity score will serve as the selection probability. 

For example, suppose we aim to fill in $\bm{q}_l$, we will compute the similarity score of current $\bm{q}_l$ with all candidate predicate embeddings $[\bm{k}_0; \bm{k}_1; \dots,  \bm{k}_{|\mathcal{X}|}]$ to find the best match to fill in. This is realized by first computing the (soft) selection score as
\begin{align}
w_{lj} = 
\frac{\exp \left( \bm{q}_l\bm{k}_j^{\top}/\tau\right)}{\sum_{j'=0, 1, \dots, |\mathcal{X}|}  \exp \left( \bm{q}_l \bm{k}_{j'}^{\top}/\tau\right)}, \quad j=0, \dots, |\mathcal{X}|
\label{Eq.3}
\end{align}
where $\tau$ is the temperature (hyperparameter) that controls the approximation error of the softmax function with the (hard) max function. Each element satisfies $ 0 < w_{lj} <1$, and each row ensures $ \sum_{j}  w_{lj}= 1$. Therefore, $w_{lj}$ can be interpreted as the selection probability of predicate $j$ to slot $l$. For each row $l$, the highest score index, denoted as $j^*(l) = \arg\max_j \{w_{lj}\}$, yields the predicate (embedding) to be selected to fill in $l$. In practice, however, we will choose to sample the best-matching predicate index according to the softmax function to introduce randomness. The additional noise may aid the rule embedding learning, preventing convergence to (very bad) suboptimal rule embeddings. This sampling from the softmax can be achieved by injecting Gumbel noise, i.e., 
\begin{align}
j^*(l) = \underset{j \in \{0, \dots, |\mathcal{X}|\}}{\operatorname{argmax}}\{\bm{q}_l\bm{k}_j^{\top}/\tau + \epsilon_j\}
\label{Eq.4}
\end{align}
where each $\epsilon_j \sim $ Gumbel (0,1). 

It is worth mentioning that $j^*(l)$ can be equal to $0$, meaning that the best-matching predicate for slot $l$ in the rule embedding is the dummy predicate embedding. By filling in the slot by dummy predicate embedding, we thereby have the flexibility to learn the rules with lengths smaller than $L$. 

We have discussed how to determine a rule formula by selecting predicate embeddings to fill in the rule embedding. Next, we will discuss how to ground the rule using data to construct the neural-symbolic feature. Let's temporarily ignore any temporal relations in each rule and consider a general static horn rule, such as
$f: \quad Y \leftarrow\left(\bigwedge_{X_u \in \mathcal{X}_f} X_u\right) 
$, where $\mathcal{X}_f$ is formed by the selected predicates and $| \mathcal{X}_f| =L$.  The neural-symbolic feature associated with this static rule can be represented as:
\begin{align}
\phi^{\text{static}}_f (\mathcal{H}_{t^-}) = \underbrace{ \prod_{l=1, \dots, L} w_{l,j^*(l)}}_{\text{similarity score}} \underbrace{\prod_{l=1, \dots, L} v_{j^*(l)}}_{\text{fact}}.
\label{eq:feature}
\end{align}
Here, each $j^*(l)$, where $l = 1, \dots, L$, is determined by sampling, and each element $0 < w_{l,j^*(l)}< 1$ is the corresponding similarity score.
$v_{j^*(l)}$, which indicates the fact, is queried from the historical events $\mathcal{H}_{t^-}$. If the corresponding event has ever occurred (i.e., the temporal predicate $X_{j^*(l)}$ has been once grounded as True), then $v_{j^*(l)}=1$; otherwise, $v_{j^*(l)}=0$.
\paragraph{Connection to Attention} Let's pause here to draw an analogy of our neural-symbolic feature construction (as shown in Eq. (\ref{eq:feature})) to the Attention mechanism~\cite{vaswani2017attention}. Recall that Attention is defined based on queries $\bm{Q} \in \mathbb{R}^{n \times d}$, keys $\bm{K} \in \mathbb{R}^{m \times d}$, and values $\bm{V} \in \mathbb{R}^{m \times v}$, and the output is computed by a weighted sum:
$$
\operatorname{Attn}(\bm{Q}, \bm{K}, \bm{V})=\operatorname{softmax}\left(\frac{\bm{Q} \bm{K}^{\top}}{\sqrt{d}}\right) \bm{V} \in \mathbb{R}^{n \times v}.
$$
We see that the way that we construct the neural-symbolic feature is similar to the attention mechanism. During the forward pass, the rule embedding (serving as query) scans across predicate embeddings (serving as keys) to find the best compositional match. Combined with observed events (serving as values), these filters produce logic-informed features.

However, our mechanism is a stricter form of attention. Instead of using all the similarity (attention) scores to compute a weighted sum output, our module approximately obtains the highest similarity score by sampling and discards all the remaining similarity weights. We use multiplication instead of summation, reflecting the nature of logic rules where all body conditions must be satisfied simultaneously for the rule to trigger. Additionally, our keys are fixed predicate embeddings with prespecified semantic meanings, which remain frozen during training.
\paragraph{Adding Temporal Relations} 
Until now, we have not taken into account any temporal relation constraints in the rule. Nevertheless, the neural-symbolic rule induction framework described above can be readily expanded to incorporate the learning of temporal relations. Building on the same idea, we can introduce a prespecified or pretrained predicate embedding to signify temporal relations ``Before'', ``Equal'', ``After'' and ``None''. This yields a matrix embedding $\bm{K}_r:=[\bm{k}_{b}; \bm{k}_{e};  \bm{k}_{a};\bm{k}_{none} ]\in  \mathbb{R}^{ 4 \times d}$. We can learn the rule embedding $\bm{Q}^r_f:=[\bm{q}_{12}; \bm{q}_{13}; \dots; \bm{q}_{L-1,L}] \in  \mathbb{R}^{\frac{L(L-1)}{2}\times d}$ to specify what temporal relation constraints should be included to the rule. Specifically, $\bm{q}_{l-1,l}$ indicates the temporal relation types of the selected predicate in slot $l-1$ and $l$. The rule embedding $\bm{Q}^r_f$ will also be filled in by the temporal predicate embedding, with the similarity scores (i.e. selection probabilities) $\bm W$ computed as Eq.~(\ref{Eq.2}). 

To determine the best-matching temporal predicate embedding to fill in the rule embedding,  similarly, one can sample an index, $\bm{j}^* \sim \text{softmax}(\bm{Q}^r_f \bm{K}_r^{\top} /\tau)$. The selected relation type is interpretable based on the sampled index for each row. The neural-symbolic feature focusing solely on temporal relations can be expressed as:
\begin{align}
\phi^{\text{temporal}}_f (\mathcal{H}_{t^-})= \underbrace{ \prod_{i,l=1, \dots, L; i<l} w_{j^*(i,l)}}_{\text{similarity score}} \underbrace{\prod_{i,l=1, \dots, L; i<l} v_{j^*(i,l)}}_{\text{fact}}.
\label{eq:feature_2}
\end{align}
Here, each fact $v_{j^*(i,l)}$ is queried from the historical events $\mathcal{H}_{t^-}$ by checking their temporal relations. Specifically, for $j^* \in \{before,equal,after,none\}$
\begin{align}
v_{j^*(i,l)}= \begin{cases}
\mathbbm{1}\left\{t_i-t_l<-\delta\right\} & j^*=before\\
\mathbbm{1}\left\{\left|t_i-t_l\right|\leq\delta\right\} & j^*=equal\\
\mathbbm{1}\left\{t_i-t_l>\delta\right\} & j^*=after \\
1 & j^*= none 
\end{cases}
\label{Eq.7}
\end{align}
Here, $\delta \geq 0$ is specified as the tolerance to accommodate data noise.  Considering the general logic rule defined in Eq. (\ref{eq:rule}), the neural-symbolic feature by combining the static and temporal parts is computed as:
\begin{align}
\phi_f (\mathcal{H}_{t^-}) = \phi^{\text{static}}_f (\mathcal{H}_{t^-}) \cdot \phi^{\text{temporal}}_f (\mathcal{H}_{t^-}) .
\label{Eq.8}
\end{align}
The calculation is to reflect that the body conditions are satisfied only when both the static part and the temporal relation part are simultaneously true. 

\subsection{More Robust Feature Construction} In the feature construction process (as detailed in Eq.~(\ref{eq:feature}), (\ref{eq:feature_2}), and (\ref{Eq.8})), the product of terms, though each close to 1, tends to decrease significantly as the number of terms increases. To maintain numerical stability, we opt for the minimum function over the product, replacing $x_1 x_2 \ldots x_N$ with $\min \left\{x_1, x_2, \ldots, x_N\right\}$. This choice ensures stability and aligns with the logical interpretation that a true rule requires each condition within it to be true. Although the minimum function is not differentiable, we address this by employing a differentiable approximation known as the soft-min function, represented as:
\begin{align}
f(x)=-\frac{1}{\rho} \log \frac{1}{N} \sum_{i=1}^N e^{-\rho x_i},
\label{Eq.9}
\end{align}
which approaches $\min _i\left|x_i\right|$ as $\rho \rightarrow+\infty$. This function is used to compute $\phi_f (\mathcal{H}_{t^-}) $,  where each $x_i$ takes values of $w$ and $v$. 
\section{Learning}
We've discussed how to construct the NS-TPP intensity using a differentiable feedforward computational graph, which allows for the learning of rules (rule embeddings $\bm{Q}$) and other continuous model parameters (such as $b_0$ and $[\gamma_f]_{f\in \mathcal{F}}$ as detailed in Eq.~(\ref{eq:intensity})) through (stochastic) gradient descent to maximize data likelihood. To learn the entire rule set, we propose a more flexible learning strategy using the sequential covering algorithm. 

This involves learning rules one by one progressively. We start with an empty set $\mathcal{F} = \emptyset$. We will learn the first rule by optimizing its rule embedding and weight using the following intensity model (constructed by a single rule) by stochastic gradient descent to maximize the likelihood:
\begin{align}
\lambda^*\left(t  \mid \mathcal{F} \right)=b_0 +  \gamma_f \phi_{f}\left(\mathcal{H}_{t^-}\right).
\label{eq:rule_1}
\end{align}
Once the optimization converges, we store the rule embedding and weight, and remove the event sequences that have been explained by this discovered rule. We update $\mathcal{F} = \{f_1\}$ and continue this process for a subsequent rule, assuming the same model as shown in Eq.~(\ref{eq:rule_1}). This procedure continues until no new rules can be added (i.e., all the event sequences have been covered). Or more often in practice we can terminate the procedure when the new discovered rule yields weight becoming smaller than some threshold. As a last step, we use the stored rule embeddings and weights to build a full model, and continue to refine the rule embeddings and weights for more accurate global model learning.

Our proposed dynamic approach eliminates the need to predefine the total number of rules, allowing the data to guide the model growing process. Additionally, we break down the overall rule problem into manageable subproblems, which simplifies the learning. 
\paragraph{Model Interpretation} For each temporal rule, the final rule formula can be directly obtained by checking the final matching score, i.e., 
\begin{equation}
 \left\{\begin{aligned} j^*(l) &= \argmax_j \{w_{lj}\}\\
 j^*(i,l) &= \argmax_{j} \{w_{j(i,l)}\} 
 \end{aligned}\right.
 \label{Eq.11}
\end{equation}
where the semantic meaning of each predicate embedding has been pre-labeled.


\section{Experiment}

\subsection{Synthetic Data Experiments}
\subsubsection{Experiment Setup}
This study utilizes a meticulously structured experimental framework that includes 30 body predicates ($X_1$ to $X_{30}$) to quantitatively assess the effectiveness of our proposed method. The framework consists of three distinct rule groups, each encompassing 1 to 3 rules to simulate varying degrees of decision logic complexity. To maintain clarity of results, each sample adheres to no more than one rule. Rule weights range from 0.40 to 1.20, indicating the differing significance of each rule within the model. The ``Ratio'' metric conveys the proportion of samples in the dataset that conform to a specific rule, offering an intuitive understanding of the rule's coverage.

Notably, samples not conforming to any rule are influenced solely by a baseline impact, ``base'', uniformly set to 0.02 across all rule groups, allowing us to control for baseline effects when assessing the model's ability to learn the importance of each rule.

The experimental datasets vary in size with 5,000, 10,000, and 20,000 instances respectively, ensuring a comprehensive evaluation of the model's performance across data scales. Results for all data sizes are presented to guarantee the integrity of the analysis and the transparency of findings. Configuration details can be found in Table \ref{tab:Table1}.
\begin{table}[ht]
\caption{\centering Ground Truth Rules and Ratios of Synthetic Dataset}
 \vskip 0.1in
\centering
\begin{tabular}{|>{\centering\arraybackslash}p{1cm}|c|c|c|}
\hline
Group & \centering Rule & Weight & Ratio \\ \hline
1 & \centering\arraybackslash \(\thead{Y \leftarrow X_1 \land X_2 \land X_3\\ \land (\text{$X_1$ before $X_2$})}\) & 0.40 & 0.20 \\ \hline
\multirow{2}{*}[-0.5em]{2} & $\thead{Y \leftarrow X_1 \land X_2 \land X_3\\ \land (\text{$X_1$ before $X_2$})}$ & 0.40 & 0.10 \\& $\thead{Y \leftarrow X_4 \land X_5\\ \land (\text{$X_4$ after $X_5$})}$ & 0.80 &  0.15 \\ \hline
\multirow{3}{*}[-1.2em]{3} & $Y$ $\leftarrow$ $X_1$ $\land$ $X_2$ $\land$ $X_3$  & 0.40 & 0.10 \\&  $\thead{Y \leftarrow X_4 \land X_5\\ \land (\text{$X_4$ after $X_5$})}$ & 0.80 & 0.15 \\ & $\thead{Y \leftarrow X_6 \land X_7\\ \land (\text{$X_6$ before $X_7$})}$ & 1.20 & 0.15 \\ \hline
\end{tabular}
\vspace{-1em}
\label{tab:Table1}
\end{table}
\subsubsection{Accuracy and Efficiency}

We conducted experiments on nine datasets with sample sizes of 5000, 10000, and 20000, corresponding to Groups 1, 2, and 3, respectively. The aim was to evaluate the accuracy and efficiency of our model.

\begin{figure}[h]
\begin{center}
\centerline{\includegraphics[width=0.9\columnwidth]{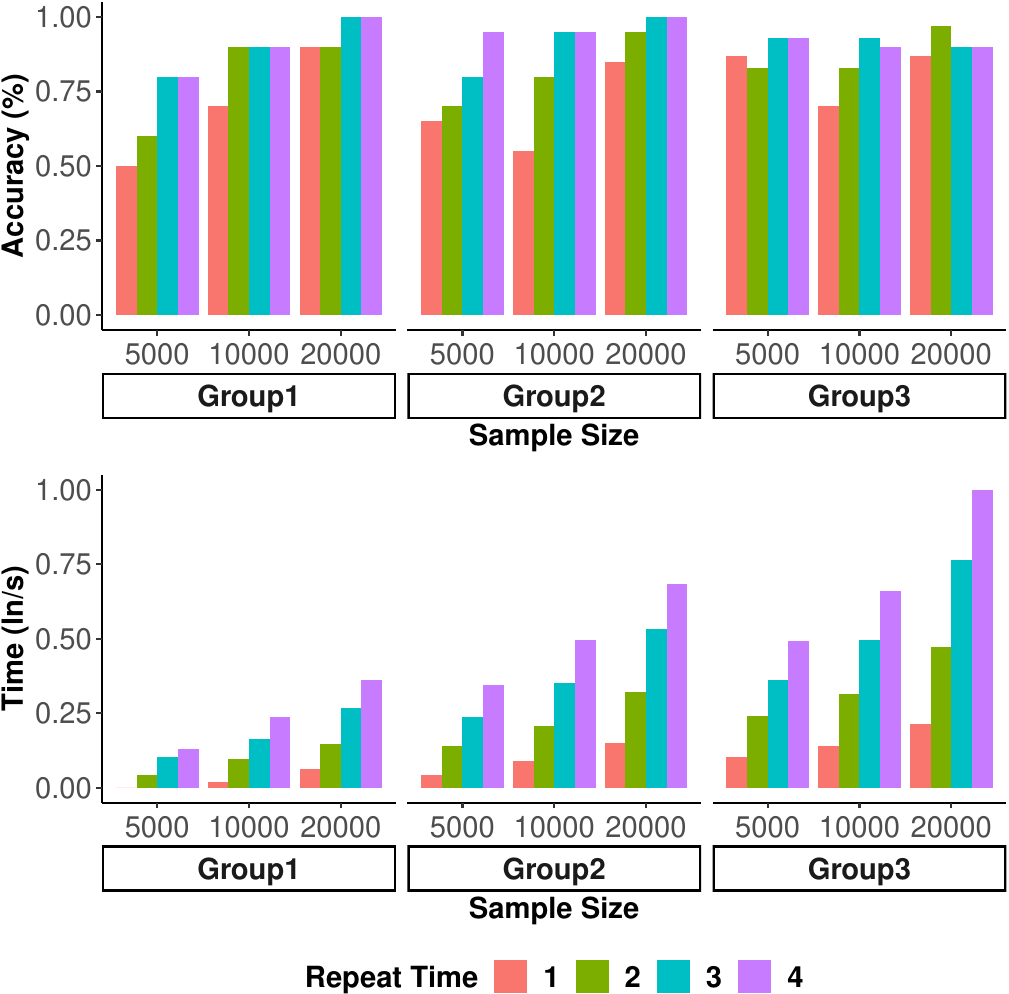}}
\vspace{-10pt}
\caption{Performance on Different Datasets at Various Repetition Times. The upper panel showcases the accuracy achieved by different groups within varying sample sizes for each repetition. The lower panel details the time efficiency across datasets, noting that time measurements, presented in seconds, have been log-normalized for clarity.}
\label{Figure2}
\end{center}
\vspace{-3em}
\end{figure}

Given the inherent randomness in rule searching, we executed multiple runs for each rule search on all datasets, varying the number of runs from 1 to 4. The rule with the minimum loss was selected as the optimal rule, ensuring consideration of different rule search iterations and identifying the top-performing logical rule. To ensure result stability and credibility, we repeated each experimental configuration ten times, reporting the average accuracy and time results in Figure \ref{Figure2}.

We must emphasize that our standards for calculating accuracy are extremely stringent; a learned rule is considered correct only if it is completely learned and aligns exactly with the Ground Truth. For instance, While correct rule is $Y\leftarrow X_1 \land X_2 \land  X_3 \land (X_1 \text{before} X_2)$, in cases of incorrect learning, we may often derive rule like $Y\leftarrow X_1 \land X_2 \land  X_3 \land (X_1 \text{before} X_2) \land (X_1 \text{before} X_3)$, which is not going too far from the ground truth.

Even under strict evaluation standards, our model demonstrates promising accuracy even with small sample sizes and a single run, with further significant improvements observed as the number of repetitions or sample size increases. Also, with increasing sample sizes or repetition counts, we observe a linear increase in time, which remains well within acceptable limits, showcasing our model's excellent scalability and practicality. It is notable that in practice, our method can perform parallel rule searches (every time we search for rules, instead of searching for a single rule, we can search for multiple rules), providing a substantial speed advantage that sets it apart from other algorithms.

\begin{figure}[h]
\begin{center}
\centerline{\includegraphics[width=0.9\columnwidth]{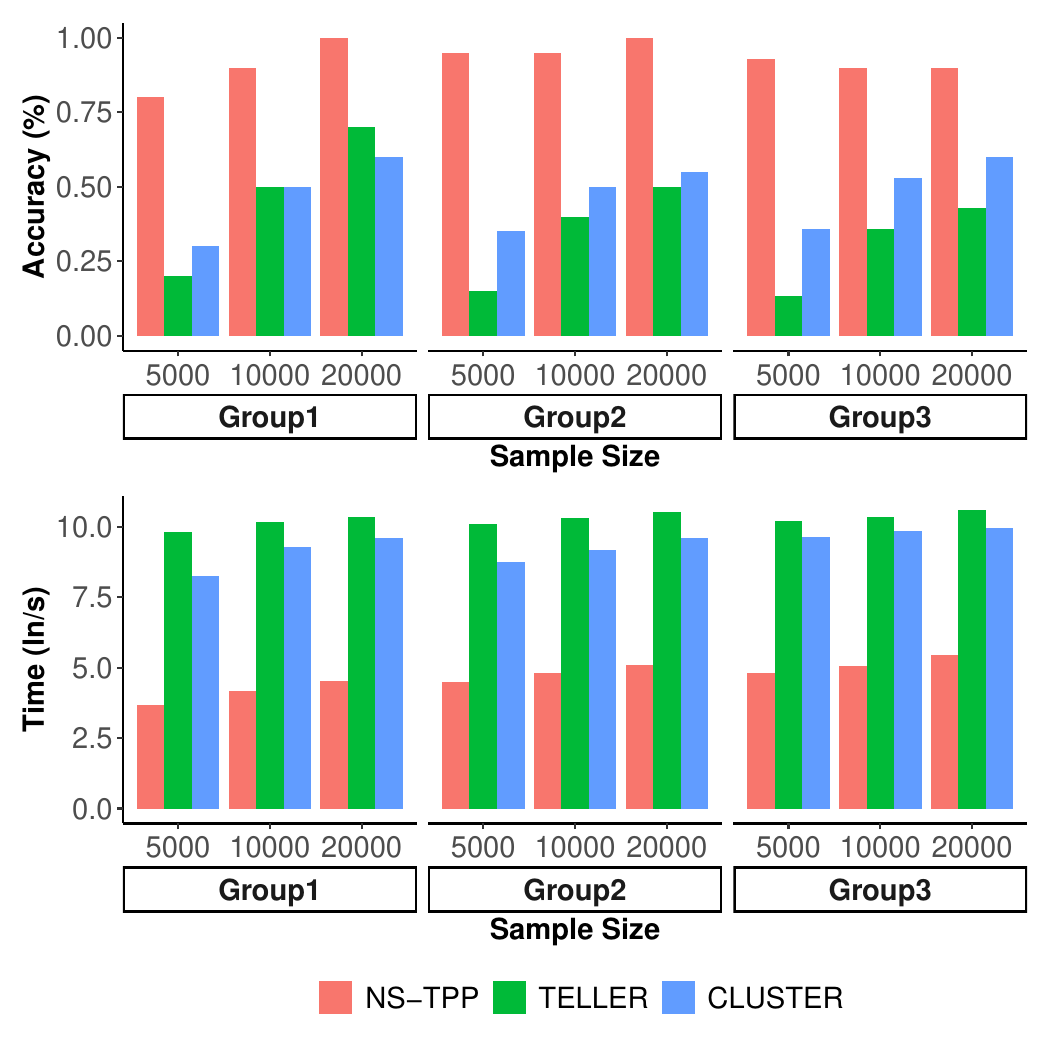}}
\vspace{-10pt}
\caption{Comparison Results of Running Time and Accuracy with TELLER and CLUSTER.}
\label{Figure3}
\end{center}
\vspace{-1em}
\end{figure}

TELLER~\cite{li2021explaining} and CLUSTER~\cite{li2024cluster} are two other algorithms capable of learning first-order temporal logic rules to explain the mechanisms behind event occurrences. We compared our method to them under the condition of searching each rule four times, and the results in terms of accuracy and time are shown in Figure \ref{Figure3}. It is evident that our algorithm significantly enhances accuracy while reducing training time when compared to the previous SOTA algorithm. On average, NS-TPP achieves a 112-fold speedup, with its accuracy significantly increasing from 49\% to 93\%.

\begin{figure}[ht]
\begin{center}
\centerline{\includegraphics[width=0.75\columnwidth]{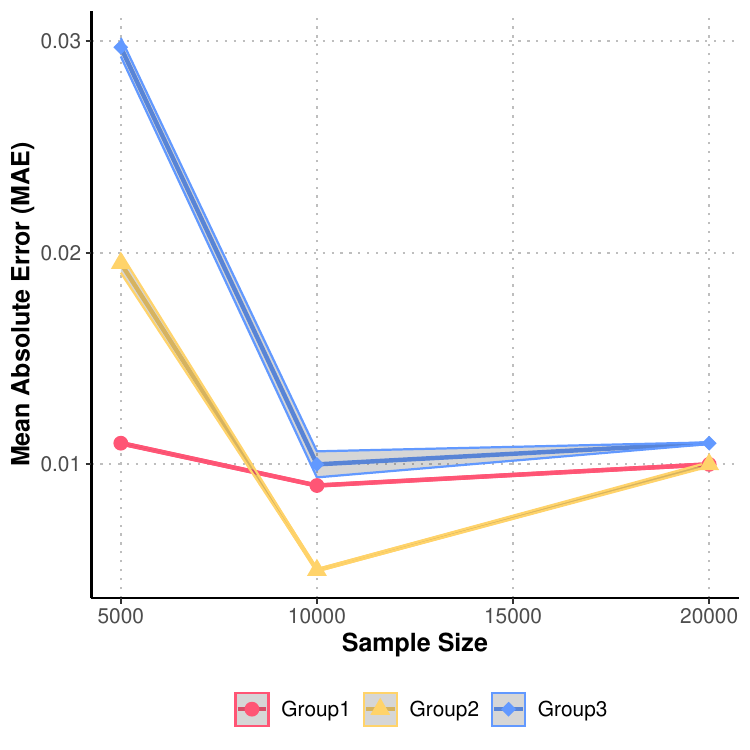}}
\vspace{-1em}
\caption{The MAE and Variance of learned rules' weight}
\label{Figure4}
\end{center}
\vspace{-2em}
\end{figure}

Building on the high accuracy of rule learning, we are able to easily obtain more precise rule weights. The Mean Absolute Error (MAE) between the rule weights calculated by our algorithm and the true values across various datasets is illustrated in Figure \ref{Figure4}.

In order to further demonstrate the superiority of our approach, we showcased the specific temporal logic rules learned by different methods on the Group-2 dataset with 10,000 samples. More result can be seen in Appendix \ref{sec:other results}. CLNN~\cite{yan2023weighted}, a method that is capable of learning fuzzy temporal logic rules, is also included in the comparison. The results of the rule learning are shown in Table \ref{Table2}. It is evident that NS-TPP can accurately learn the rules along with their corresponding weights, whereas other baseline methods encounter difficulties in the rule-learning phase.
\begin{table}[h]
\caption{Learned rules and corresponding weights on the Group-2 dataset under different models.For each model, we report the best result of four runs as evaluated by log-likelihood.}
\vskip 0.1in
\centering
\small
\begin{tabular}{|>{\centering\arraybackslash}m{1.2cm}|c|>{\centering\arraybackslash} p{1cm}|}
\hline
Model & Learned Rules (Group2) & \multicolumn{1}{c|}{Weight} 
\\ \hline
\multirow{2}{*}[-0.2em]{$\thead{\text{Ground}\\ \text{Truth}}$} & $\thead{Y \leftarrow X_1 \land X_2 \land X_3 \land (X_1~\text{before}~X_2)}$ & \multicolumn{1}{c|}{0.40} \\ & $\thead{Y \leftarrow X_4 \land X_5 \land (X_4~\text{after}~X_5)}$ & \multicolumn{1}{c|}{0.80} \\ \hline
\multirow{2}{*}[-0.3em]{NS-TPP} & $\thead{Y \leftarrow X_1 \land X_2 \land X_3 \land (X_1~\text{before}~X_2)}$ &  \multicolumn{1}{c|}{0.40} \\ & $\thead{Y \leftarrow X_4 \land X_5  \land (X_4~\text{after}~X_5)}$ & \multicolumn{1}{c|}{0.79}\\ \hline
\multirow{2}{*}[-0.4em]{CLUSTER} & $\thead{Y \leftarrow X_2 \land X_4 \land X_5 \land (X_5~\text{before}~X_4)}$ & \multicolumn{1}{c|}{0.58} \\ & $\thead{Y \leftarrow X_1 }$ & \multicolumn{1}{c|}{0.40} \\ \hline
\multirow{2}{*}[-0.2em]{TELLER} & $\thead{Y \leftarrow X_4 \land X_5 \land (X_4~\text{after}~X_5)}$ & \multicolumn{1}{c|}{0.76} \\ & $\thead{Y \leftarrow X_5}$ & \multicolumn{1}{c|}{0.73} \\ \hline
\multirow{2}{*}[-1em]{CLNN} & $\thead{Y \leftarrow (X_3\text{after}X_{13})^{2.79}\\ \land (X_{15}\text{after}X_{21})^{2.71} \land (X_{24})^{2.46} }$ & \multicolumn{1}{c|}{0.87} \\ & $\thead{Y \leftarrow (X_1 \text{before} X_{18})^{4.17}\\ \land(X_1\text{after}X_9)^{2.21} \land (X_9\text{before}X_{17})^{2.17} }$ & \multicolumn{1}{c|}{0.01} \\ 
\hline
\end{tabular}
\label{Table2}
\end{table}
\normalsize
\subsubsection{Event Prediction}
In addition to the aforementioned baseline methods capable of learning temporal logic rules, we also compared our model with some neural network-based methods specialized in event prediction, forecasting the occurrence of target events, and using MAE as the evaluation metric for the event prediction task. For baseline descriptions and environment configuration, refer to Appendices \ref{sec:baselines} and \ref{sec:configuration}. As shown in Table {\color{blue}\ref{Table3}}, our model consistently excels in all metrics, matching or surpassing the baseline performance.
\vspace{-5pt}
\begin{table}[h]
\vspace{-1em}
\caption{Event time prediction MAE on synthetic dataset (Group-3, 20000 samples)}
\vskip 0.0in
\label{Table:MAE}
\begin{center}
\begin{small}
\begin{sc}
\begin{tabular}{l|c}
\hline
\textbf{Method} & $\thead{\textbf{Group-3}\\ (\text{20000 samples})}$ \\
\hline
THP~\cite{zuo2020transformer}     &26.1545    \\
RMTPP~\cite{du2016recurrent} & 31.0179 \\
ERPP~\cite{xiao2017modeling}  &28.8209   \\
GCH~\cite{xu2016learning}   &26.7682    \\
LG-NPP~\cite{zhang2021learning}   &32.9013  \\
GM-NLF~\cite{eichler2017graphical}  &26.8176   \\
TELLER~\cite{li2021explaining}   & 27.8301   \\
CLNN~\cite{yan2023weighted}  & 29.7430   \\
CLUSTER~\cite{li2024cluster} &26.0351        \\
\textbf{NS-TPP} & \textbf{24.8616}  \\
\hline
\end{tabular}
\label{Table3}
\end{sc}
\end{small}
\end{center}
\vspace{-2em}
\end{table}

\subsection{Real Data Experiments}
\subsubsection{Experiment Setup}

Our research involved the study of two datasets: the Car-Following dataset for assessing autonomous vehicle behavior, and the LowUrine dataset, which encompasses a wealth of medical records from ICU patients.

Within the Car-Following dataset, we gleaned five key driving behavior features from over 460 hours of driving data, leading to the documentation of 10,042 sequences. Our endeavor in this dataset is to analyze these sequences to mine for vehicle-following patterns and to deduce the underlying temporal logic rules that govern such dynamics.

The LowUrine dataset, derived from the MIMIC-IV\footnote{\url{https://mimic.mit.edu/}}, focuses on the electronic health records of $4074$ ICU patients diagnosed with sepsis, capturing the physiological changes that occur leading up to the critical juncture of septic shock. A thorough analysis was conducted on 29 vital signs and laboratory tests, selected based on recommendations from previous validated studies~\cite{komorowski2018artificial}. Special attention was given to recording the first abnormal values within the 48 hours prior to an abnormal urine output event. The analysis of this dataset aims to identify early warning signals and reveal logical patterns that may indicate the onset of septic shock, offering practical significance for clinical intervention. Details on data processing can be found in Appendix \ref{sec:variables describe}.

\subsubsection{Discovered Logic Rules}
In the Car-Following dataset, we explored temporal logic rules influencing vehicle dynamics by sequentially treating different events as the target event. While this dataset is relatively simpler, it is still crucial for understanding vehicle behavior patterns. In contrast, the LowUrine dataset is more complex and of greater importance, where our focus was on mining rules leading to sudden abnormal decreases in urine output. Urine output, being a significant health status indicator, especially when low urine output may signal impending septic shock, is critical for monitoring in ICU settings. Therefore, in this dataset, particular attention was paid to instances where urine output becomes abnormal after maintaining normal levels for at least 48 hours, as these events are more meaningful for prediction and explanation.

In Table {\color{blue}\ref{Table4}}, we showcase a selection of key logic rules discovered using our methodology, along with their corresponding weights. Notably, for the medical logic rules identified within the LowUrine dataset, our findings align with conclusions from various existing studies and are substantiated by a wealth of medical literature. For a detailed discussion of how these literatures corroborate our findings, refer to Appendix \ref{sec:medical ref}.
\vspace{-15pt}
\begin{table}[h]
\caption{Learned Rule\protect\footnotemark for different Dataset.}
\vskip 0.1in
\centering
\begin{threeparttable} 
\begin{tabular}{|c|c|c|}
\hline
\multicolumn{1}{|c|}{Dataset} & \multicolumn{1}{c|}{Rule} & \multicolumn{1}{c|}{Weight} \\ \hline
\multirow{3}{*}{Car-Following} & $ \thead{\text{A} \leftarrow \text{C}} $ & 0.58  \\ \cline{2-3} 
 & $\text{C} \leftarrow \text{Fa}$ & 0.66 \\ \cline{2-3} 
 & $\thead{\text{F} \leftarrow \text{A}  \land \text{D}  \land (\text{A}~\text{before}~\text{D} )}$ & 0.62 \\ \hline
\multirow{6}{*}{\text{LowUrine}} & $\text{LowUrine} \leftarrow
\text{VO2P}$ & 0.37 \\ \cline{2-3} 
 &  $ \text{LowUrine} \leftarrow
\text{RRate} \land \text{He}$ & 0.49 \\ \cline{2-3} 
 &  $\text{LowUrine} \leftarrow \text{BUN} \land  \text{LA}$ & 0.41 \\ \cline{2-3} 
 & $\thead{\text{LowUrine} \leftarrow
\text{RRate}\land \text{LA}\\ \land (\text{RRate}~\text{after}~\text{LA} )}$ & 0.32 \\ \cline{2-3} 
 & $\thead{\text{LowUrine} \leftarrow \text{Ma} \land \text{VO2P} \land \text{LA}\\ \land (\text{Ma}~\text{before}~\text{VO2P})\\ \land (\text{Ma}~\text{before}~\text{LA})\\ \land (\text{VO2P}~\text{before}~\text{LA})}$ &  0.38 \\ \hline
\end{tabular}
\end{threeparttable}
\label{Table4}
\end{table}
\footnotetext{For detailed explanations of the predicates across all rules, refer to Appendix \ref{sec:glossary}.}
\subsubsection{Event Prediction}

In our experimental section, we employed the same baseline models as in the synthetic data experiments, using Mean Absolute Error (MAE) as the evaluation metric to predict ``Low Urine'' events in the LowUrine dataset and ``Constant Speed Following'' events in the Car-Following dataset. The performance of our model across these two datasets is presented in Table \ref{Table5}. The results indicate that our model outperforms all the baseline models in predicting both types of events.
\begin{table}[h]
\caption{Event time prediction MAE on real datasets}
\label{sample-table}
\begin{center}
\begin{small}
\begin{sc}
\begin{tabular}{l|cc}
\hline
\textbf{Method} & \textbf{Car-Following} & \textbf{LowUrine} \\
\hline
THP      &3.8920   & 2.4234\\
RMTPPP    &4.5575   & 2.4643\\
ERPP     &4.0947   & 2.6122\\
GCH      &3.9819   & 2.5367\\
LG-NPP   &4.2787   & 2.5672\\
GM-NLFF   &4.7195   & 2.6925\\
TELLER   &4.6012   & 2.4401\\
CLNN      &4.3842    &2.4371  \\
CLUSTER    &3.7255   &2.3675 \\
\textbf{NS-TPP} & \textbf{3.1614} &\textbf{2.3262} \\
\hline
\end{tabular}
\label{Table5}
\end{sc}
\vspace{-4em}
\end{small}
\end{center}

\end{table}

\section{Conclusion}
In this paper, we introduce a new approach that integrates neural-symbolic rule induction with temporal point process models, focused on efficiently mining temporal logic rules to better understand anomalies in complex event sequences. This method not only enhances the efficiency of the rule learning process but also ensures the interpretability of the results. Extensive testing on both synthetic and real datasets has revealed significant advantages of this approach in terms of efficiency and accuracy in rule mining, demonstrating its practicality and effectiveness in complex data analysis.

\section*{Acknowledgements}
Shuang Li’s research was in part supported by the National Science and Technology Major Project under grant No. 2022ZD0116004, the NSFC under grant No. 62206236, Shenzhen Science and Technology Program under grant No. JCYJ20210324120011032,   Shenzhen Key Lab of Cross-Modal Cognitive Computing under grant No. ZDSYS20230626091302006, and Guangdong Key Lab of Mathematical Foundations for Artificial Intelligence.

\section*{Impact Statement}
Our research introduces a novel neuro-symbolic framework for temporal logic induction, marking a significant advancement in machine learning's capability to process and interpret complex temporal data. By seamlessly integrating neural networks with symbolic reasoning, our approach not only enhances model interpretability but also improves predictive accuracy across diverse datasets. This work opens new avenues for developing AI systems that can better understand and predict temporal sequences, with broad implications for fields such as autonomous systems, healthcare monitoring, and financial forecasting. Our framework's flexibility and efficiency showcase its potential to foster AI solutions that not only mimic human behavior and cognition but also enhance decision-making with ethical and transparent attributes.

\bibliography{example_paper}
\bibliographystyle{icml2024}

\newpage
\appendix
\onecolumn
\setcounter{table}{0}
\renewcommand\thetable{\thesection.\arabic{table}}
\section{Result On Other Datasets}
\label{sec:other results}
\begin{table}[h]
\caption{Learned rules and corresponding weights on the Group-1 dataset under different models. For each model, we report the best result of four runs as evaluated by log-likelihood.}
\vskip 0.1in
\centering
\small
\begin{tabular}{|>{\centering\arraybackslash}m{1.2cm}|c|>{\centering\arraybackslash} p{0.6cm}|}
\hline
Model & Learned Rules (Group1) & \multicolumn{1}{c|}{Weight} 
\\ \hline
\multirow{1}{*}[0.9em]{$\thead{\text{Ground}\\ \text{Truth}}$} & $\thead{Y \leftarrow X_1 \land X_2 \land X_3 \land (X_1~\text{before}~X_2)}$ & \multicolumn{1}{c|}{0.40} \\ \hline
\multirow{1}{*}[-0.3em]{NS-TPP} & $\thead{Y \leftarrow X_1 \land X_2 \land X_3 \land (X_1~\text{before}~X_2)}$ &  \multicolumn{1}{c|}{0.41} \\ \hline
\multirow{1}{*}[-0.4em]{CLUSTER} & $\thead{Y \leftarrow X_1 \land X_2 \land X_3 \land (X_1~\text{before}~X_2)}$ & \multicolumn{1}{c|}{0.37} \\ \hline
\multirow{1}{*}[-0.2em]{TELLER} & $\thead{Y \leftarrow X_1}$ & \multicolumn{1}{c|}{0.68} \\ \hline
\multirow{1}{*}[-0.2em]{CLNN} & $\thead{Y \leftarrow X_{27}^{0.981} \land X_{3}^{0.484} \land X_{9}^{0.215}}$ & \multicolumn{1}{c|}{0.29} \\ 
\hline
\end{tabular}
\label{Table6}
\end{table}

\begin{table}[h]
\caption{Learned rules and corresponding weights on the Group-3 dataset under different models. For each model, we report the best result of four runs as evaluated by log-likelihood.}
\vskip 0.1in
\centering
\small
\begin{tabular}{|>{\centering\arraybackslash}m{1.2cm}|c|>{\centering\arraybackslash} p{0.6cm}|}
\hline
Model & Learned Rules (Group3) & \multicolumn{1}{c|}{Weight} 
\\ \hline
\multirow{3}{*}[-0.4em]{$\thead{\text{Ground}\\ \text{Truth}}$} & $\thead{Y \leftarrow X_1 \land X_2 \land X_3}$ & \multicolumn{1}{c|}{0.40} \\
& $\thead{Y \leftarrow X_4 \land X_5 \land (X_4~\text{after}~X_5)}$ & \multicolumn{1}{c|}{0.80} \\
& $\thead{Y \leftarrow X_6 \land X_7 \land (X_6~\text{before}~X_7)}$ & \multicolumn{1}{c|}{1.20} \\ \hline
\multirow{3}{*}[-0.3em]{NS-TPP} & $\thead{Y \leftarrow X_1 \land X_2 \land X_3}$ & \multicolumn{1}{c|}{0.39} \\
& $\thead{Y \leftarrow X_4 \land X_5 \land (X_4~\text{after}~X_5)}$ & \multicolumn{1}{c|}{0.79} \\
& $\thead{Y \leftarrow X_6 \land X_7 \land (X_6~\text{before}~X_7)}$ & \multicolumn{1}{c|}{1.20} \\ \hline
\multirow{3}{*}[-0.4em]{CLUSTER} & $\thead{Y \leftarrow X_{16} \land X_{23} \land X_{24}}$ & \multicolumn{1}{c|}{0.46} \\
& $\thead{Y \leftarrow X_1 \land X_2 \land X_3 \land (X_1~\text{before}~X_2)}$ & \multicolumn{1}{c|}{0.47} \\
& $\thead{Y \leftarrow X_{12}}$ & \multicolumn{1}{c|}{0.55} \\ \hline
\multirow{3}{*}[-0.2em]{TELLER} & $\thead{Y \leftarrow X_7}$ & \multicolumn{1}{c|}{1.28} \\
& $\thead{Y \leftarrow X_6 \land X_7}$ & \multicolumn{1}{c|}{1.24} \\
& $\thead{Y \leftarrow X_6}$ & \multicolumn{1}{c|}{1.11} \\ \hline
\multirow{3}{*}[-2em]{CLNN} & $\thead{Y \leftarrow (X_1 \text{before} X_{25})^{1.02} \land X_3^{0.740} \land X_2^{0.571}}$ & \multicolumn{1}{c|}{0.58} \\
& $\thead{Y \leftarrow X_2^{0.791} \land X_7^{0.491} \land X_{18}^{0.347}}$ & \multicolumn{1}{c|}{0.13} \\
& $\thead{Y \leftarrow (X_{20} \text{before} X_8)^{0.631} \land X_4^{0.61} \land (X_{13} \text{before} X_{10})^{0.417}}$ & \multicolumn{1}{c|}{0.62} \\
\hline
\end{tabular}
\label{Table7}
\end{table}

\section{MIMIC-IV Dataset Preprocessing Details}
\label{sec:variables describe}

MIMIC-IV\footnote{\url{https://mimic.mit.edu/}} is a publicly available database sourced from the electronic health record of the Beth Israel Deaconess Medical Center~\citep{johnson2023mimic}. The information available includes patient measurements, orders, diagnoses, procedures, treatments, and deidentified free-text clinical notes. Sepsis is a leading cause of mortality in the ICU, particularly when it progresses to septic shock. Septic shocks are critical medical emergencies, and timely recognition and treatment are crucial for improving survival rates. In the real-world experiments on the MIMIC-IV dataset, we aim to find logic rules related to septic shocks for the whole patient samples and infer the most likely rule reasons for specific patients, which would be potential early alarms when some abnormal indicators occur.

\paragraph{Patients} We select 4074 patients that satisfied the following criteria from the dataset: (1) The patients are diagnosed with sepsis~\citep{saria2018individualized}. (2) Patients, if diagnosed with sepsis, the timestamps of any clinical testing, specific lab values, timestamps of medication administration, and corresponding dosage were not missing.

\paragraph{Outcome} Real-time urine output was treated as the outcome indicator since low urine output signals directly indicate a poor circulatory system and is a warning sign of septic shock.

\paragraph{Data Preprocessing} In our experiment, we focus on the electronic health records of 4,074 ICU patients diagnosed with sepsis, capturing the physiological changes that occur leading up to the critical juncture of septic shock. A thorough analysis was conducted on 29 vital signs and laboratory tests, selected based on recommendations from previous validated studies~\cite{komorowski2018artificial}. Special attention was given to recording the first abnormal values within the 48 hours prior to an abnormal urine output event. The analysis of this dataset aims to identify early warning signals and reveal logical patterns that may indicate the onset of septic shock, offering practical significance for clinical intervention.

These risk factors are commonly assessed in sepsis patients to monitor their clinical status and guide appropriate interventions. The interpretation of these factors requires clinical judgment and consideration of the patient's overall condition. Appendix C shows the categories of some variables extracted from the MIMIC-IV dataset and their reference range.

\section{About Baselines}
\label{sec:baselines}
We consider the following baselines through synthetic data experiments and healthcare data experiments to compare the rule learning ability and event prediction with our proposed model:
\paragraph{Neural-based (black-box) models for irregular event data}

\begin{itemize}
\item Transformer Hawkes Process (THP)~\citep{zuo2020transformer}: It is a sophisticated model that combines the Transformer's sequence modeling capabilities with the Hawkes process for handling irregularly timed events. This innovative approach allows for effective forecasting and understanding of complex temporal event dependencies.
\item Recurrent Marked Temporal Point Processes (RMTPP) ~\citep{du2016recurrent}: It is a model that utilizes recurrent neural networks to analyze and predict the timing and types of events in sequences. It excels in handling complex temporal relationships in data, making it valuable for applications requiring a detailed understanding of event sequences and their dynamics.
\item ERPP~\citep{xiao2017modeling}: It is a neural network approach for modeling event sequences, focusing on capturing the complex temporal patterns and dependencies between events. This model is notable for its ability to effectively handle a wide range of event-based datasets, providing insights into the underlying structure and dynamics of temporal data.
\item LG-NPP algorithm ~\citep{zhang2021learning}: It is an innovative neural process model designed for learning and predicting the intricate patterns in event sequences. This algorithm stands out for its effectiveness in capturing the long-term dependencies and subtle nuances within sequential data, making it highly applicable in complex temporal analysis tasks.
\end{itemize}

\paragraph{Simple parametric/nonparametric models for irregular event data}
\begin{itemize}
\item Granger Causal Hawkes (GCH) ~\citep{xu2016learning}: It is a statistical approach that combines Granger causality analysis with the Hawkes process to understand the influence of past events on future occurrences. It excels in identifying causal relationships in temporal data, making it particularly useful in fields where understanding the impact of past events on future dynamics is crucial.
\item GM-NLF algorithm ~\citep{eichler2017graphical}: This is a sophisticated algorithm designed for analyzing complex nonlinear relationships in time series data. It is particularly notable for its ability to model and predict intricate patterns and dependencies, enhancing the understanding of dynamic systems in various domains.
\end{itemize}

\paragraph{Logical models for irregular event data}
\begin{itemize}
\item Clock Logic Neural Networks (CLNN) ~\citep{yan2023weighted}: It represents a novel approach in neural network design, integrating time-aware mechanisms to better handle temporal data. This model is particularly effective in capturing both the sequential and timing aspects of events, offering enhanced performance in tasks requiring precise temporal understanding and prediction.
\item TELLER~\citep{li2021explaining}: This is a cutting-edge neural network model designed for temporal and event-based data analysis. It stands out for its ability to intricately model and predict complex patterns in sequential data, making it highly effective in applications requiring deep temporal understanding and forecasting.
\item CLUSTER~\cite{li2024cluster}: This is an automated method for uncovering “if-then” logic rules to explain observational events. This approach demonstrates accurate performance in both discovering rules and identifying root causes.
\end{itemize}

We compared our model with some models from previous studies on the same dataset, finding that not only does it run in a shorter time, but it also achieves higher accuracy.

\section{Experimental Environment Configuration}
\label{sec:configuration}
For our proposed method, all experiments were conducted on a Linux server with an Intel(R) Xeon(R) Gold 6248R CPU @ 3.00GHz and 30Gi of memory, running Ubuntu 20.04.5 LTS. Due to the modest size of our model parameters, CPU execution was found to be more efficient. Hence, while all baseline methods except TELLER were run on GPU, we opted to perform our experiments on the CPU. The coding environment utilized was Python 3.9.12, with PyTorch 2.0.1 serving as the primary machine-learning framework.
\section{Glossary}\label{sec:glossary}
\subsection{Car-Following Dataset}
\begin{center}
\begin{tabular}{c|c}
\hline Predicates & Explanation \\
\hline Fa & Free acceleration \\
\hline C & Cruising at a desired speed \\
\hline A & Acceleration following a leading vehicle \\
\hline D & Deceleration following a leading vehicle \\
\hline F & Constant speed following \\
\hline
\end{tabular}
\end{center}
\subsection{LowUrine Dataset} 
The 29 extracted predicates can be categorized into the following five groups:

\begin{itemize}
    \item Vital Signs:
    \begin{itemize}
        \item Heart Rate: The number of times the heart beats per minute. An elevated or abnormal heart rate may indicate physiological stress or an underlying condition.
        \item Arterial Blood Pressure (systolic, mean, diastolic): Measures the force exerted by the blood against the arterial walls during different phases of the cardiac cycle. Abnormal blood pressure values may indicate cardiovascular dysfunction or organ perfusion issues.
        \item Temperature (Celsius): Body temperature is a measure of the body's internal heat. Abnormal temperatures may indicate infection, inflammation, or other systemic disorders.
        \item Respiratory Rate(RRate): The number of breaths taken per minute. Abnormal respiratory rates may suggest respiratory distress or dysfunction.
        \item $SpO_2$: Oxygen saturation level in the blood. Decreased $SpO_2$ levels may indicate inadequate oxygenation.
    \end{itemize}
    
    \item Biochemical Parameters:
    \begin{itemize}
        \item Potassium, Sodium, Chloride, Glucose: Electrolytes and blood sugar levels that help maintain essential bodily functions. Abnormal levels may indicate electrolyte imbalances, metabolic disorders, or organ dysfunction.
        \item Blood Urea Nitrogen (BUN), Creatinine: Indicators of renal function. Elevated levels may suggest impaired kidney function.
        \item Magnesium(Ma), Ionized Calcium: Important minerals involved in various physiological processes. Abnormal levels may indicate electrolyte imbalances or organ dysfunction.
        \item Total Bilirubin: A byproduct of red blood cell breakdown. Elevated levels may indicate liver dysfunction.
        \item Albumin: A protein produced by the liver. Abnormal levels may indicate malnutrition, liver disease, or kidney dysfunction.
    \end{itemize}

    \item Hematological Parameters
    \begin{itemize}
        \item Hemoglobin(He): A protein in red blood cells that carries oxygen. Abnormal levels may indicate anemia or oxygen-carrying capacity issues.
        \item White Blood Cell (WBC): Cells of the immune system involved in fighting infections. Abnormal levels may indicate infection or inflammation.
        \item Platelet Count: Blood cells responsible for clotting. Abnormal levels may suggest bleeding disorders or impaired clotting ability.
        \item Partial Thromboplastin Time (PTT), Prothrombin Time (PT), INR: Tests that assess blood clotting function. Abnormal results may indicate bleeding disorders or coagulation abnormalities.
    \end{itemize}

    \item Blood Gas Analysis
    \begin{itemize}
        \item pH (Arterial): A measure of blood acidity or alkalinity. Abnormal pH values may indicate acid-base imbalances or respiratory/metabolic disorders.
        \item Arterial Base Excess: Measures the amount of excess or deficit of base in arterial blood. Abnormal levels may indicate acid-base imbalances or metabolic disturbances.
        \item Arterial CO2 Pressure(AO2P), Venous O2 Pressure(VO2P): Parameters that assess respiratory and metabolic function. Abnormal values may indicate respiratory failure or metabolic disturbances.
    \end{itemize}

    \item Metabolic Parameter
    \begin{itemize}
        \item Lactic Acid(LA): An indicator of tissue perfusion and oxygenation. Elevated levels may suggest tissue hypoxia or impaired cellular metabolism.
    \end{itemize}

\end{itemize}
\section{Medical References}\label{sec:medical ref}
In our clinical research employing the MIMIC-IV dataset, we strengthened our findings with corroborative evidence from medical literature, demonstrating the robustness and clinical applicability of our methodology. This integrative process ensures our discovered rules not only align with expert insights but are also grounded in established medical knowledge, enhancing the interpretability and real-world applicability of temporal logic rules in healthcare analytics.
\begin{itemize}
    \item \textbf{Rule 1: LowUrine $\leftarrow$ VO2P}: These rules involve venous O2 pressure, it linked to cardiac output and tissue hypoxia in septic shock ~\citep{Mohsenin2017Practical,Rhodes2004Early}.

    \item \textbf{Rule 2: LowUrine $\leftarrow$ RRate $\land$ He}: The studies indicate that effective management of respiratory function and maintaining adequate hemoglobin levels are crucial for ensuring efficient oxygen delivery and preventing complications like low urine output. This highlights the interconnectedness of respiratory health, oxygen transport capacity, and kidney function in maintaining overall systemic health~\citep{Kallet2009The,Aprilia2022Hubungan}.

    \item \textbf{Rule 3: LowUrine $\leftarrow$ BUN $\land$ LA}:Research highlights the significant impact of metabolic disturbances, such as hyperuricemia and the risk of lactic acidosis from medications like metformin, on renal function and urine output. Managing these conditions through urinary alkalization and careful medication management is crucial for preventing renal complications and maintaining adequate urine output~\citep{Shekarriz2002Uric,Inzucchi2014Metformin}.

    \item \textbf{Rule 4: LowUrine $\leftarrow$ RRate $\land$ LA $\land$ (RRate after LA)}: Abnormal levels of lactate are typically induced by tissue hypoxia or metabolic disturbances, while subsequent abnormalities in respiratory rate may represent the body's compensatory effort to eliminate excess acid metabolites through respiration. Together, these symptoms may indicate a deteriorating clinical condition, progressing towards sepsis~\citep{suetrong2016lactic}.
    
    \item \textbf{Rule 5: LowUrine $\leftarrow$ Ma $\land$ VO2P $\land$ LA $\land$ (Ma before VO2P) $\land$ (Ma before LA) $\land$ (VO2P before LA)}: Research indicates that hypomagnesemia is associated with increased cardiovascular risk, which may indirectly impact kidney function and urine output~\citep{Wei2006Relationship}. Additionally, inadequate oxygen delivery and elevated lactate levels signal systemic hypoperfusion, including renal hypoperfusion, potentially leading to reduced urine output~\citep{Landow1993Splanchnic}. These findings underscore the importance of magnesium levels, venous O2 pressure, and lactate in maintaining kidney health and appropriate urine output.
\end{itemize}

\end{document}


\section{About Baselines}
\label{sec:baselines}
We consider the following baselines through synthetic data experiments and healthcare data experiments to compare the rule learning ability and event prediction with our proposed model:
\paragraph{Neural-based (black-box) models for irregular event data}

\begin{itemize}
\item Transformer Hawkes Process (THP)~\citep{zuo2020transformer}: It is a sophisticated model that combines the Transformer's sequence modeling capabilities with the Hawkes process for handling irregularly timed events. This innovative approach allows for effective forecasting and understanding of complex temporal event dependencies.
\item Recurrent Marked Temporal Point Processes (RMTPP) ~\citep{du2016recurrent}: It is a model that utilizes recurrent neural networks to analyze and predict the timing and types of events in sequences. It excels in handling complex temporal relationships in data, making it valuable for applications requiring detailed understanding of event sequences and their dynamics.
\item ERPP~\citep{xiao2017modeling}: It  is a neural network approach for modeling event sequences, focusing on capturing the complex temporal patterns and dependencies between events. This model is notable for its ability to effectively handle a wide range of event-based datasets, providing insights into the underlying structure and dynamics of temporal data.
\item LG-NPP algorithm ~\citep{zhang2021learning}: It is an innovative neural process model designed for learning and predicting the intricate patterns in event sequences. This algorithm stands out for its effectiveness in capturing the long-term dependencies and subtle nuances within sequential data, making it highly applicable in complex temporal analysis tasks.
\end{itemize}

\paragraph{Simple parametric/nonparametric models for irregular event data}
\begin{itemize}
\item Granger Causal Hawkes (GCH) ~\citep{xu2016learning}: It is a statistical approach that combines Granger causality analysis with the Hawkes process to understand the influence of past events on future occurrences. It excels in identifying causal relationships in temporal data, making it particularly useful in fields where understanding the impact of past events on future dynamics is crucial.
\item GM-NLF algorithm ~\citep{eichler2017graphical}: This is a sophisticated algorithm designed for analyzing complex nonlinear relationships in time series data. It is particularly notable for its ability to model and predict intricate patterns and dependencies, enhancing the understanding of dynamic systems in various domains.
\end{itemize}

\paragraph{Logical models for irregular event data}
\begin{itemize}
\item Clock Logic Neural Networks (CLNN) ~\citep{yan2023weighted}: It represents a novel approach in neural network design, integrating time-aware mechanisms to better handle temporal data. This model is particularly effective in capturing both the sequential and timing aspects of events, offering enhanced performance in tasks requiring precise temporal understanding and prediction.
\item TELLER ~\citep{li2021explaining}: This is a cutting-edge neural network model designed for temporal and event-based data analysis. It stands out for its ability to intricately model and predict complex patterns in sequential data, making it highly effective in applications requiring deep temporal understanding and forecasting.
\item Cluster ~\citep{li2021explaining}: This is  an automated method for uncovering “if-then” logic rules to explain observational events. This approach demonstrates accurate performance in both discovering rules and identifying root causes.
\end{itemize}

We compared our model with some models from previous studies on the same dataset, finding that not only does it run in a shorter time, but it also achieves higher accuracy.

\paragraph{Itemset mining methods}
\begin{itemize}
\item Apriori ~\citep{agrawal1994fast}: It is a fundamental algorithm used for mining frequent itemsets in large databases and for discovering association rules. It is renowned for its efficiency and simplicity, making it a foundational tool in the field of data mining for uncovering relationships between seemingly unrelated data.
\item NEclatcloesed ~\citep{aryabarzan2021neclatclosed}: It is an advanced approach in the domain of frequent itemset mining. It is designed to efficiently discover closed itemsets, offering improved performance and scalability compared to traditional methods, particularly in large and complex datasets.
\end{itemize}

\paragraph{Sequential pattern mining}
\begin{itemize}
\item CM-spade ~\citep{fournier2014fast}: This is an innovative algorithm for sequential pattern mining. It is designed for efficiently discovering frequent, sequential patterns in large datasets, and is particularly noted for its effectiveness in handling massive and complex data sequences.
\item VGEN ~\citep{fournier2014vgen}: This is a significant algorithm in the field of sequential pattern mining. It is designed for the efficient extraction of vertical sequential patterns in large datasets, demonstrating a high level of effectiveness and scalability in processing complex sequences of data.
\end{itemize}

\paragraph{Sequential rule mining}
\begin{itemize}
\item ERMiner ~\citep{fournier2014erminer}: This is a powerful algorithm for mining sequential rules in large databases. It is known for its efficiency in discovering meaningful patterns and rules within sequence data, making it a valuable tool in data analysis, especially in environments with extensive and complex datasets.
\end{itemize}

\paragraph{Post-hoc method}
\begin{itemize}
\item Dynamask ~\citep{crabbe2021explaining}: It is an innovative algorithm designed for interpretability in machine learning. It focuses on dynamically uncovering the importance of different features in neural network decisions, offering enhanced transparency and understanding in model predictions, particularly useful in complex decision-making scenarios.
\end{itemize}

\section{MIMIC-IV dataset preprocessing and risk factors extracting}
\label{sec:variables describe}

MIMIC-IV\footnote{\url{https://mimic.mit.edu/}} is a publicly available database sourced from the electronic health record of the Beth Israel Deaconess Medical Center~\citep{johnson2023mimic}. Information available includes patient measurements, orders, diagnoses, procedures, treatments, and deidentified free-text clinical notes. Sepsis is a leading cause of mortality in the ICU, particularly when it progresses to septic shock. Septic shocks are critical medical emergencies, and timely recognition and treatment are crucial for improving survival rates. In the real-world experiments on MIMIC-IV dataset, we aim to find logic rules related to septic shocks for the whole patient samples and infer the most likely rule reasons for specific patients, which would be potential early alarm when some abnormal indicators occur.

\paragraph{Patients} We select 4074 patients that satisfied the following criteria from the dataset: (1) The patients are diagnosed with sepsis~\citep{saria2018individualized}. (2) Patients, if diagonized with sepsis, the timestamps of any clinical testing, specific lab values, timestamps of medication administration and corresponding dosage were not missing.

\paragraph{Outcome} Real time urine output was treated as the outcome indicator since low urine output signals directly indicate a poor circulatory system and is a warning sign of septic shock. 

\paragraph{Risk Factors} Suggested by \citep{komorowski2018artificial}, we extract 28 risk factors associated with sepsis which are consistent with expert consensus. Based on the distinct clinical characteristics of these risk factors, they can be categorized into the following five groups:

\begin{itemize}
    \item Vital Signs:
    \begin{itemize}
        \item Heart Rate: The number of times the heart beats per minute. An elevated or abnormal heart rate may indicate physiological stress or an underlying condition.
        \item Arterial Blood Pressure (systolic, mean, diastolic): Measures the force exerted by the blood against the arterial walls during different phases of the cardiac cycle. Abnormal blood pressure values may indicate cardiovascular dysfunction or organ perfusion issues.
        \item Temperature (Celsius): Body temperature is a measure of the body's internal heat. Abnormal temperatures may indicate infection, inflammation, or other systemic disorders.
        \item Respiratory Rate: The number of breaths taken per minute. Abnormal respiratory rates may suggest respiratory distress or dysfunction.
        \item $SpO_2$: Oxygen saturation level in the blood. Decreased $SpO_2$ levels may indicate inadequate oxygenation.
    \end{itemize}
    
    \item Biochemical Parameters:
    \begin{itemize}
        \item Potassium, Sodium, Chloride, Glucose: Electrolytes and blood sugar levels that help maintain essential bodily functions. Abnormal levels may indicate electrolyte imbalances, metabolic disorders, or organ dysfunction.
        \item Blood Urea Nitrogen (BUN), Creatinine: Indicators of renal function. Elevated levels may suggest impaired kidney function.
        \item Magnesium, Ionized Calcium: Important minerals involved in various physiological processes. Abnormal levels may indicate electrolyte imbalances or organ dysfunction.
        \item Total Bilirubin: A byproduct of red blood cell breakdown. Elevated levels may indicate liver dysfunction.
        \item Albumin: A protein produced by the liver. Abnormal levels may indicate malnutrition, liver disease, or kidney dysfunction.
    \end{itemize}

    \item Hematological Parameters
    \begin{itemize}
        \item Hemoglobin: A protein in red blood cells that carries oxygen. Abnormal levels may indicate anemia or oxygen-carrying capacity issues.
        \item White Blood Cell (WBC): Cells of the immune system involved in fighting infections. Abnormal levels may indicate infection or inflammation.
        \item Platelet Count: Blood cells responsible for clotting. Abnormal levels may suggest bleeding disorders or impaired clotting ability.
        \item Partial Thromboplastin Time (PTT), Prothrombin time (PT), INR: Tests that assess blood clotting function. Abnormal results may indicate bleeding disorders or coagulation abnormalities.
    \end{itemize}

    \item Blood Gas Analysis
    \begin{itemize}
        \item pH (Arterial): A measure of blood acidity or alkalinity. Abnormal pH values may indicate acid-base imbalances or respiratory/metabolic disorders.
        \item Arterial Base Excess: Measures the amount of excess or deficit of base in arterial blood. Abnormal levels may indicate acid-base imbalances or metabolic disturbances.
        \item Arterial CO2 Pressure, Venous O2 Pressure: Parameters that assess respiratory and metabolic function. Abnormal values may indicate respiratory failure or metabolic disturbances.
    \end{itemize}

    \item Metabolic Parameter
    \begin{itemize}
        \item Lactic Acid: An indicator of tissue perfusion and oxygenation. Elevated levels may suggest tissue hypoxia or impaired cellular metabolism.
    \end{itemize}

\end{itemize}

These risk factors are commonly assessed in sepsis patients to monitor their clinical status and guide appropriate interventions. The interpretation of these factors requires clinical judgment and consideration of the patient's overall condition. Tab.~\ref{Variables description} shows the categories of the 28 variables extracted from MIMIC-IV dataset, and their reference range.

\paragraph{Data Preprocessing} Due to the frequent fluctuations in urine output within the ICU setting, we considered only those instances in which urine output became abnormal after maintaining a normal level for at least 48 hours. These instances were regarded as valid target events that hold significance for prediction and explanation. Additionally, for each patient, we selectively extracted the initial time period that met the criteria. Regarding the risk factors, we documented the time points at which these variables first became abnormal from normal within the 48-hour period preceding the transition of urine output from normal to abnormal.

\section{Detailed Analysis for Synthetic Data Experiments}
\paragraph{Setup} In our empirical analysis, we systematically explore various aspects to evaluate the efficiency and performance of our method: \textbf{(i) Diverse Ground Truth Rules:} We vary the number of ground truth rules, ranging from 1 to 3, to examine the model’s performance across different complexity levels.\textbf{(ii) Variation in Data Size:} We conduct experiments with varying data sizes, including sequences of 5000, 10000 and 20000 instances, to investigate our proposed EM algorithm’s scalability under different data scales.
\paragraph{Implementation Details} 
The E-step, which refers to inference, has a closed-form.The M-step is more involved. In M-step, we learn model parameters, including the predicate assignment to rules, the temporal relationships among the selected predicates, the impact weight of each rule, and the prior distribution of the rules, using coordinate descent. In other words, we optimize one component of the model parameters by holding the others. 
\paragraph{Results Analysis} 
We conducted experiments on two datasets with different groups (varying in the number of learning rules) under three sample sizes: 5000, 10000, and 20000. These experiments were compared with various models. The rules we learned showed a significantly higher accuracy rate, and our algorithm, based on our matching principles, also significantly improved the learning speed of the model compared to other models.